\documentclass[10pt,twocolumn,letterpaper]{article}

\usepackage[final]{cvpr}      
\usepackage{colortbl}
\usepackage{wrapfig}
\usepackage{adjustbox}
\usepackage{colortbl}
\usepackage{multirow}
\usepackage{url}            
\usepackage{booktabs}       
\usepackage{amsfonts}       
\usepackage{nicefrac}       
\usepackage{microtype}      
\usepackage{xcolor}         
\usepackage{wrapfig}
\usepackage{multirow}
\usepackage{lipsum} 
\usepackage{graphicx}
\usepackage{subcaption}

\usepackage{amsmath}
\usepackage{amssymb}
\usepackage{mathtools}
\usepackage{booktabs}
\usepackage{amsthm}
\usepackage{wasysym}
\usepackage{makecell}
\usepackage{changepage,threeparttable}
\usepackage{bbm}
\usepackage{amssymb}
\definecolor{cvprblue}{rgb}{0.21,0.49,0.74}
\usepackage[pagebackref,breaklinks,colorlinks,allcolors=cvprblue]{hyperref}

\def\paperID{****} 
\def\confName{CVPR}
\def\confYear{2025}

\title{Tra-MoE: Learning Trajectory Prediction Model from Multiple Domains \\ for Adaptive Policy Conditioning}

\author{Jiange Yang $^{1, 2}$ \quad Haoyi Zhu$^{2, 3}$ \quad Yating Wang$^{2, 4}$  \quad Gangshan Wu$^{1}$ \quad Tong He$^{2}$ \quad Limin Wang$^{1, 2}$\thanks{Corresponding author.}\\
$^1$State Key Laboratory for Novel Software Technology, Nanjing University, China \\$^2$Shanghai AI Laboratory, 
$^3$University of Science and Technology of China,  $^4$Tongji University \\
\tt\small jiangeyang.jgy@gmail.com, \{gswu,lmwang\}@nju.edu.cn, \\ 
\tt\small \{zhuhaoyi, wangyating, hetong\}@pjlab.org.cn \\
\textbf{\normalsize\url{https://github.com/MCG-NJU/Tra-MoE}}
}

\begin{document}
\maketitle

\begin{abstract}
  Learning from multiple domains is a primary factor that influences the generalization of a single unified robot system. In this paper, we aim to learn the trajectory prediction model by using broad out-of-domain data to improve its performance and generalization ability. Trajectory model is designed to predict any-point trajectories in the current frame given an instruction and can provide detailed control guidance for robotic policy learning. To handle the diverse out-of-domain data distribution, we propose a sparsely-gated MoE (\textbf{Top-1} gating strategy) architecture for trajectory model, coined as \textbf{Tra-MoE}. The sparse activation design enables good balance between parameter cooperation and specialization, effectively benefiting from large-scale out-of-domain data while maintaining constant FLOPs per token. In addition, we further introduce an adaptive policy conditioning technique by learning 2D mask representations for predicted trajectories, which is explicitly aligned with image observations to guide action prediction more flexibly. We perform extensive experiments on both simulation and real-world scenarios to verify the effectiveness of Tra-MoE and adaptive policy conditioning technique. We also conduct a comprehensive empirical study to train Tra-MoE, demonstrating that our Tra-MoE consistently exhibits superior performance compared to the dense baseline model, even when the latter is scaled to match Tra-MoE's parameter count.
\end{abstract}

\vspace{-6mm}



\section{Introduction}
\label{sec:intro}

\vspace{-0.5mm}


\begin{figure}
    \centering
    \includegraphics[width=0.80\linewidth]{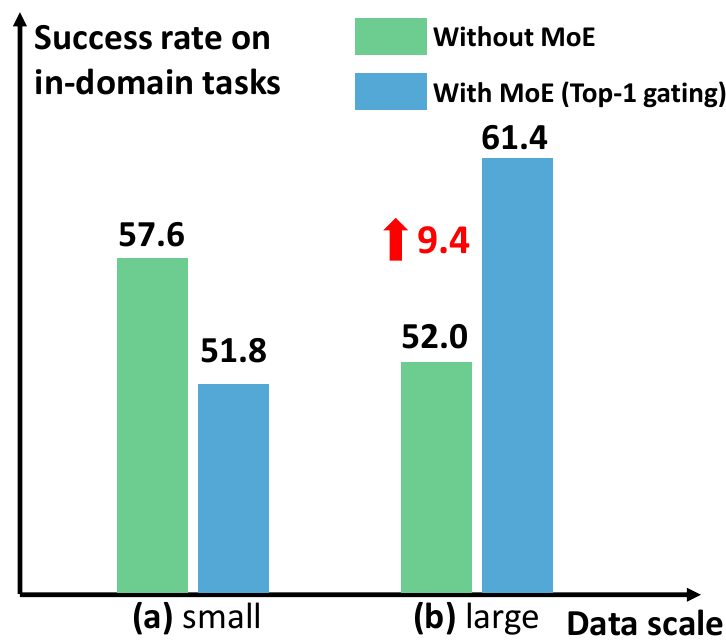}
    \caption{\textbf{(a)} Training with small-scale, in-domain data. \textbf{(b)} Joint training with in-domain data and large-scale, out-of-domain data.}
    \label{vis}
    \vspace{-6mm}
\end{figure}

In computer vision and natural language processing, significant progress has been made through learning models from multiple domains data. For example, many works use board data to perform self-supervised pre-training, and then adapt to diverse downstream tasks through zero-shot~\citep{gpt,clip,gpt4} or fine-tuning~\citep{bert,mocov3,mae,videomae}. Additionally, some research efforts~\citep{unified,unipe,unipemoe,gato} focus on learning a generalist model capable of unifying various tasks. In contrast, in robot learning, due to data scarcity and homogeneity, some groundbreaking methods~\citep{e2e,hand} resort to training only using in-domain data. Recently, some works~\citep{afp,affordance, mimicplay,plan,lapa} have shifted to learn from action-free human video data. A particularly scalable and promising paradigm~\citep{robotap,atm,flow,track2act,flip} involves learning a trajectory prediction model from action-free video data, and then using small-scale action-labeled demonstrations to learn trajectory-guided policies, thereby achieving greater sample efficiency and better generalization.

\begin{figure}[htbp]
\centering
\includegraphics[width=0.99\linewidth]{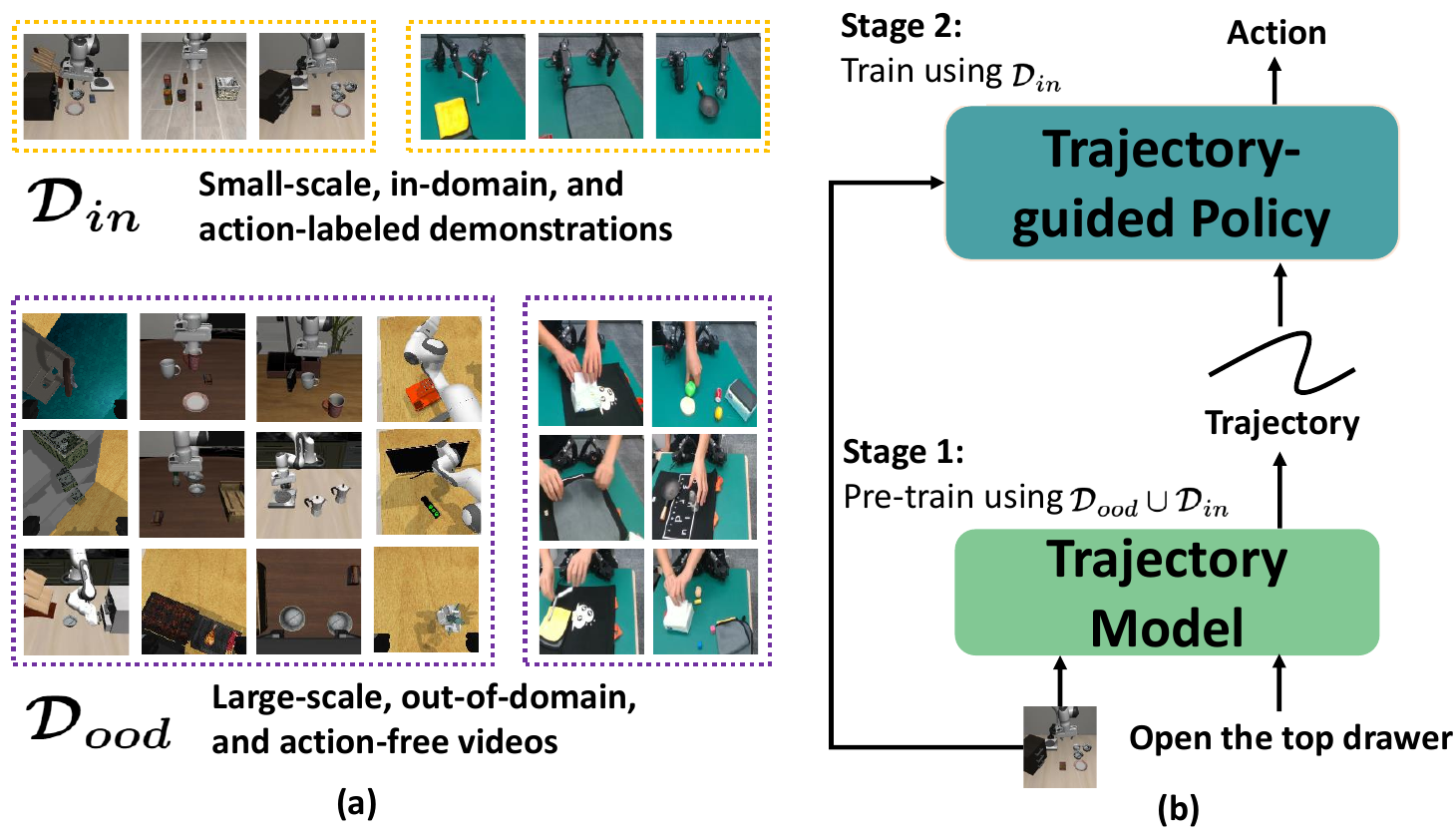}
\vspace{-0.4em}
\caption{{\bf Training trajectory prediction model from multiple domains}. \textbf{(a)} The visualization of Dataset $\boldsymbol{\mathcal{D}_{ood}}$ and $\boldsymbol{\mathcal{D}_{in}}$. The former may contain additional environments, objects, skills and embodiments. \textbf{(b)} Our pipeline: first co-training trajectory prediction model and then adapting it for downstream policy learning.}
\vspace{-6mm}
\label{setup}
\end{figure}

Despite the potential benefits, it still remains underexplored how to effectively leverage broad out-of-domain video data for jointly training trajectory models. In particular, the impact of incorporating data encompassing diverse environments, objects, skills and embodiments has not fully investigated for in-domain tasks. For instance, ATM~\citep{atm}, a notable work in this field, mainly relies on in-domain data (same tasks and environments) for training both trajectory models and policies. 
We argue that learning from multiple domains is the primary factor to build a powerful trajectory model for policy guidance.
Therefore, we propose to train trajectory model under the setup of jointly learning from large-scale out-of-domain and small-scale in-domain data, as shown in Fig.~\ref{setup}(a). However, directly using large-scale out-of-domain video data for hybrid pre-training involves two critical challenges: \textcolor{blue}{\textbf{(\romannumeral 1)}} Joint training on data from different environments, objects, skills, and embodiments will add learning difficulty and might lead to optimization conflicts. It is unclear how to effectively fuse the complementary information of multiple domains data within a single unified model. \textit{Naively expanding the out-of-domain training data does not necessarily lead to improved performance on in-domain tasks}, as shown in Fig.~\ref{vis}, which results in a 5.6 (57.6 → 52.0) performance decrease on in-domain tasks. \textcolor{blue}{\textbf{(\romannumeral 2)}} How to ensure both high performance and computational efficiency during trajectory model scaling up process.


To address these challenges, we design a new sparsely-gated Mixture-of-Expert (MoE) architecture~\citep{hardmoe} to scale up our trajectory model, coined as {\bf Tra-MoE}. Specifically, we integrate several MoE blocks to replace the original transformer blocks. This design enables efficient joint training of most parameters on the broad out-of-domain data, capturing the complementary patterns for mutual cooperation and the shared knowledge for better generalization. Meanwhile, for different data and different tokens within the same data, our Tra-MoE naturally forms a specialization when activating different experts. To maintain high computational efficiency while scaling up model capacity, we implement a \textbf{top-1} gating strategy for token-choice, ensuring constant FLOPs per token. In sparse MoE, some auxiliary losses are commonly used to enhance performance, including the router z-loss~\citep{stmoe} for improving training stability and the load-balancing loss~\citep{shard} for balancing expert activations. We further conduct a comprehensive empirical study on scaling up trajectory models and find that the touter z-loss improves performance while the load-balancing loss tends to cause performance degradation. Furthermore, we observe that when expanding pre-training data from small-scale in-domain to large-scale in-domain and out-of-domain data, employing MoE (Tra-MoE) can lead to obvious performance improvement on in-domain tasks, whereas the dense counterpart (Tra-baseline) suffers from performance drop, as illustrated in Fig.~\ref{vis}. Notably, even when expanding the dense Tra-baseline's parameter to match that of Tra-MoE, performance improvements remain elusive. These findings collectively demonstrate the effectiveness of our sparsely-gated MoE architecture, attributing its capability of balancing parameter cooperation and specialization when expanding training with out-of-domain data.


Furthermore, given a pre-trained trajectory model, how to effectively condition the policy module using the 2D trajectories still remain an unresolved challenge. We further propose an adaptive policy conditioning technique for trajectory-guided policy. This enables better spatial alignment between 2D trajectories and images, while achieving more flexible trajectory representation and guidance for the policy. Finally, we also conduct extensive real-world experiments to further strengthen our conclusions. In summary, our main contributions can be summarized as follows:


\begin{itemize}
    \item We propose a sparsely-gated MoE architecture to learn trajectory models, called as Tra-MoE, with large-scale, out-of-domain, and action-free video data, coupled with a comprehensive empirical study on its training techniques.
    \item Extensive simulation and real-world environments demonstrate that our Tra-MoE (with Top-1 gating) can more effectively benefit from large-scale out-of-domain data (such as cross-embodiments data and different physics engine simulated data) compared to the dense baseline.
    \item We demonstrate that Tra-MoE still significantly outperform expanded dense counterpart with equivalent parameters, highlighting the effectiveness of sparse MoE architecture in leveraging out-of-domain data through better parameter cooperation and specialization.
    \item We propose an adaptive policy conditioning technique for trajectory-guided policy, which ensures adaptive 2D trajectory representation and explicit spatial alignment with the observations, thereby achieving superior performance.
\end{itemize}


    
   




\section{Related Work}

\label{sec:related-work}
\noindent \textbf{Learning from multiple domains in Robot Learning.} Scaling laws~\citep{laws} have fueled substantial advancements in various fields. In robot learning, many recent studies~\citep{gato,vima,rt1,robocat} investigate how compute, model size, and training data quantity affect the model performance on robotic manipulation tasks. Based on this, recent works have focused on collecting larger-scale real-world robotic datasets~\citep{rh20t,bridgedata,droid,robomind,agibot}, and have successfully trained several robotics foundational models~\citep{rt2,robofla,openx,octo,openvla,levine,hpt,pi0} that demonstrate superior transfer capabilities. It is worth noting that these works mainly employ pre-training and fine-tuning paradigm rather than joint co-training paradigm, while focusing on evaluating the combination generalization and neglecting the performance impact on in-domain tasks.  Additionally, these works directly combine heterogeneous robotic data with different action and observation spaces for joint training, potentially leading to sub-optimal solutions. In contrast, we leverage sparsely-gated MoE to incorporate larger-scale, out-of-domain, and action-less video data. Furthermore, our experiments demonstrate the importance of MoE for expanding out-of-domain data due to its superior capacity for parameter cooperation and specialization.



\noindent \textbf{Mixture-of-Experts.} The Mixture-of-Experts (MoE) framework~\citep{moesurvey} is built upon a simple yet effective concept: dividing a model into specialized components, or `experts', each focusing on distinct tasks or data aspects. This paradigm allows for selective activation of relevant experts based on the input, thereby maintaining computational efficiency while leveraging a vast reservoir of specialized knowledge through increased model capacity. Recently, in computer vision, natural language processing, and multi-modal, there has been a surge of foundation models trained based on the MoE architecture, including Swin-Moe~\citep{swinmoe}, MoE-LLaVA~\citep{Moe-llava}, DeepSeek-V2~\citep{deepseek}, Mixtral-8x22B~\citep{Mixtral-8x22B}, and so on. MoE typically encompass dense~\citep{softmoe} and sparse~\citep{squad} variants. Dense MoE activates all experts in each iteration, while sparse MoE activates only a subset, generally resulting in lower computational overhead. However, sparse MoE can suffer from expert load imbalance and training instability issues. Techniques such as adding noise, load balancing loss~\citep{shard}, and router z-loss~\citep{stmoe} are often employed to mitigate these issues and improve performance. In our work, we conduct a comprehensive empirical study of these techniques when training our sparsely-gated MoE-based trajectory model.

\begin{figure*}[h]
\centering
\includegraphics[width=0.87\linewidth]{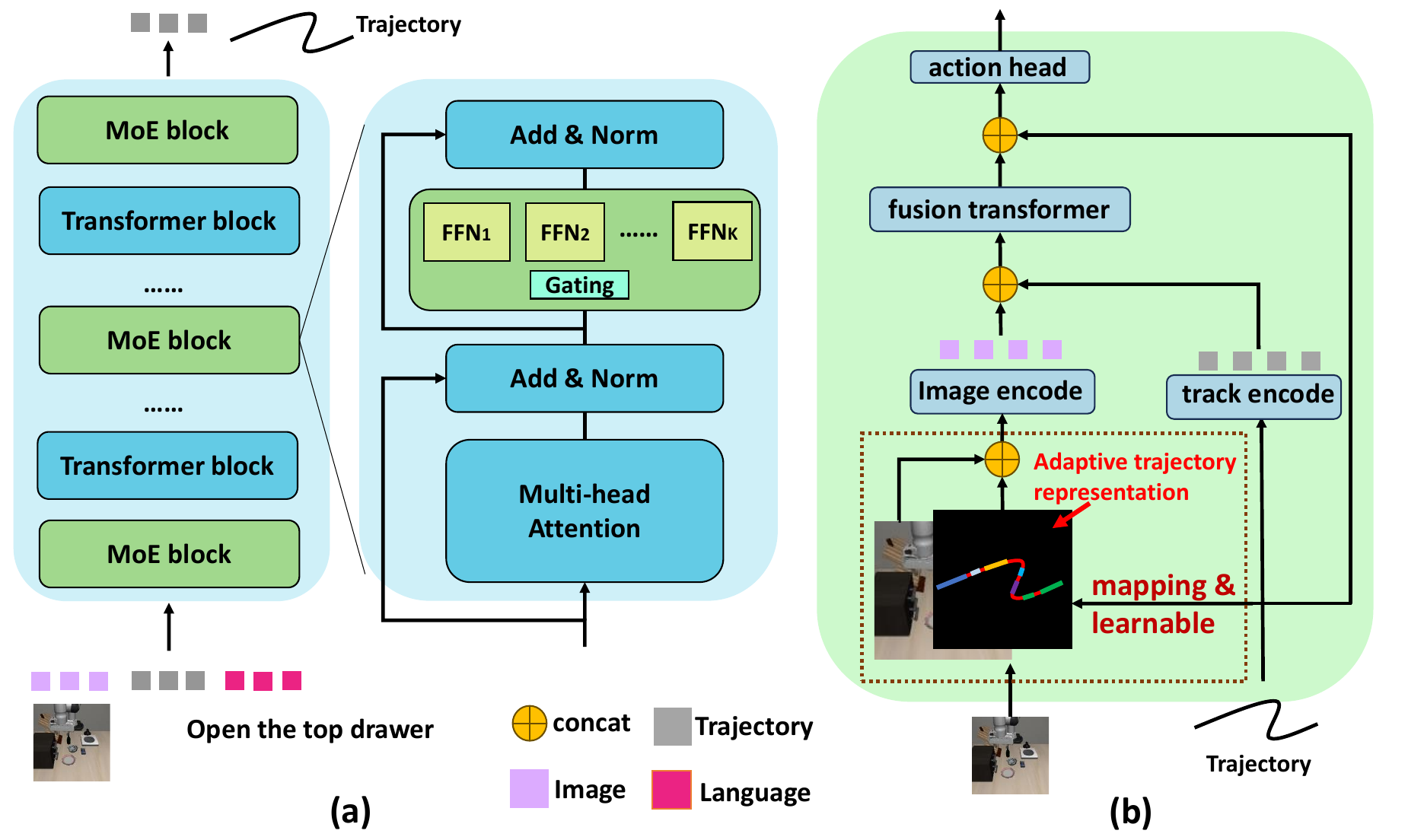}
\caption{ \textbf{(a)} The pipeline of our sparsely-gated MoE-based trajectory model (\textbf{Tra-MoE}). \textbf{(b)} The pipeline of our trajectory-guided policy using the adaptive policy conditioning technique. Mapping means concatenating the trajectory mask with image observations, while learnable refers to setting each point in the trajectory mask as a learnable embedding.}
\vspace{-5mm}
\label{MoE}
\end{figure*}


\noindent \textbf{Representations and Techniques for Policy Conditioning.} 
Policy conditioning equips a multi-task robot policy with the ability to modulate its behavior by incorporating task-specific guidance information. This information, which may take various forms, guides the policy's action generation, enabling a single model to effectively learn and perform diverse tasks. Classic policy conditioning representations often encompass task identifiers~\citep{taskid}, natural language instructions~\citep{shridhar2022cliport,hiveformer,table,rt1}, goal images~\citep{goalimage}, and demonstration videos~\citep{bcz,vid2robot}. These are typically integrated with visual observations within the policy through concatenation or attention mechanisms. Recent works have explored alternative policy conditioning representations to enhance generalization and sample efficiency. These include leveraging internet-trained models to obtain bounding box~\citep{moo} and object mask~\citep{zhu,pave,pave2,moka}, as well as learning plan~\citep{plan}, affordance~\citep{affordance,RT-Affordance}, future observations~\citep{du1,du2,stp,cor}, 2D or 3D trajectory~\citep{mimicplay,atm,generalflow,track2act,flow} from human video data, and also hand-drawn trajectory~\citep{rt-tra} or sketch~\citep{sketch}. Different from these approaches that directly utilize raw or manually crafted signals for integration with visual observations, we introduce an adaptive policy conditioning technique that explicitly maps 2D trajectories onto the visual observations in space and encodes them as learnable embeddings. These designs ensure adaptive 2D trajectory representation and explicit spatial alignment with the visual observations.

\section{Methodology}

In this section, we first describe our problem formulation in Sec.~\ref{Problem Formulation}. Subsequently, we give an overview of our proposed framework, which is composed of a sparsely-gated MoE-based trajectory prediction model and a trajectory-guided policy model using the adaptive policy conditioning technique, detailed in Sec.~\ref{Architecture}. Finally, we describe our training process and loss functions in Sec.~\ref{Training}.

\subsection{Problem Formulation}
\label{Problem Formulation}

Following the setup of previous works~\citep{atm,track2act,flow}, our approach encompasses a trajectory prediction model to conduct language-conditioned any-point trajectory prediction and a trajectory-guided policy model to predict executable robot actions. In particular, as depicted in Fig.~\ref{setup}, we consider two types of datasets for training, which are large-scale, out-of-domain, and action-free videos $\boldsymbol{\mathcal{D}_{ood}}$, as well as small-scale, in-domain, and action-labeled demonstrations $\boldsymbol{\mathcal{D}_{in}}$. For trajectory prediction model, we employ $\boldsymbol{\mathcal{D}_{ood}}$ and $\boldsymbol{\mathcal{D}_{in}}$ for joint pre-training, whereas for trajectory-guided policy model, we only use $\boldsymbol{\mathcal{D}_{in}}$ for training.

\subsection{Architecture}
\label{Architecture}

\noindent \textbf{ATM Revisited.} In this paper, we use ATM~\cite{atm} as the baseline for our method. The overall pipeline of ATM is illustrated on Fig.~\ref{setup}(b). It first learns a trajectory model from video data to perform language-conditioned trajectory prediction. The predicted trajectories is then used to learn a trajectory-guided policy using robot demonstrations. Specifically, as for trajectory model, its objective is to predict the future of any-point trajectories in a frame. Formally, given an image observation $o_t$ at timestep $t$, any set of 2D query points simpled on the image frame $\mathbf{p}_t = \{p_{t, k}\}_{k=1}^K$, and a language instruction $\ell$, the trajectory model learns a mapping $\mathbf{p}_{t:t+H} = \tau_\theta(o_t, \mathbf{p}_t, \ell)$ that predicts the coordinates of these query points in the camera frame for the next $H$ time steps, where $\mathbf{p}_{t:t+H} \in \mathbb{R}^{H \times K \times 2}$. For the trajectory-guided policy, it leverages the current observation $o_t$ and the predicted trajectory $\textbf{p}_{t:t+H}$ to predict the subsequent robot actions.


\noindent \textbf{Sparsely-gated MoE-based trajectory model (Tra-MoE).} ATM proposes a track transformer to implement the trajectory model. In track transformer, language tokens, image tokens, and track tokens to be decoded involve global interactions within the transformer block. However, directly processing diverse domains and multiple modalities data within a unified transformer may yield suboptimal results. Hence, we propose our Tra-MoE framework. To develop our sparsely-gated MoE-based trajectory model, as depicted in Fig.~\ref{MoE}(a), we replace certain layers of the transformer with MoE blocks. In each MoE block, it incorporates multiple feedforward networks (FFNs), each designated as an expert, and utilizes a gating function to activate a selected subset (Top-$K$) of these experts. In this paper, we set $K$ to a fixed value of $1$ to ensure that the FLOPs per token remain constant relative to the baseline. Formally, the gating network $\mathcal{G}$, parameterized by $\mathbf{\Theta}$ and typically consisting of a linear-softmax network, yields the output $\mathcal{G}(\mathbf{x_s}; \mathbf{\Theta})$, where $\mathbf{x_s}$ represents our token sequences input. We formulate the sparsely-gated token-choice mechanism as follows:

\vspace{-3mm}

\begin{align}
    &\mathcal{G}(\mathbf{x_s}; \mathbf{\Theta})_i = \operatorname{softmax}(\operatorname{Top-K}(g(\mathbf{x_s}; \mathbf{\Theta}) , k))_i,\label{eq:sparse_gating}\\
    &\operatorname{Top-K}(g(\mathbf{x_s}; \mathbf{\Theta}), k)_i = 
    \begin{cases}
    g(\mathbf{x}; \mathbf{\Theta})_i, &\text{cond}, \\
    -\infty, &\text{oth}.
    \end{cases},\\
    &\mathrm{cond}:\text{if $g(\mathbf{x}; \mathbf{\Theta})_i$ is in top-$k$ elements of $g(\mathbf{x}; \mathbf{\Theta})$}.
    \label{eq:topk}
\end{align}

\vspace{-3mm}

More specifically, $\operatorname{Top-K}(\cdot, k)$ function retains only the top-$k$ values in a vector, setting the rest to $-\infty$. Consequently, the output of the sparsely-gated MoE layer can be formulated as:
\begin{align}
&\mathcal{F}_{\mathrm{sparse}}^{\mathrm{MoE}}(\mathbf{x_s}; \mathbf{\Theta}, \{\mathbf{W}_i\}_{i=1}^{K}) =
\sum_{i=1}^{K}\mathcal{G}(\mathbf{x_s}; \mathbf{\Theta})_i f_{i}(\mathbf{x_s}; \mathbf{W}_i),\label{sparse_MoE}
\end{align}
where $f_{i}$ represents the i-th expert, usually a linear-GELU-linear FFN layer, is parameterized by $\mathbf{W}_i$.






\noindent \textbf{Trajectory-guided policy using the adaptive policy conditioning technique.} After completing the trajectory model learning, we develop our trajectory-guided policy through the adaptive policy conditioning technique based on the initial ATM, as shown in Fig.~\ref{MoE}(b). In our policy model, independently encoded track tokens and image tokens undergo a early fusion within the fusion transformer. Subsequently, the fused features are further integrated with the raw trajectory data through a late fusion mechanism along the channel dimension, ensuring the preservation of original trajectory information for more accurate action prediction. 

Fundamentally, the policy model bridges the gap between the predicted 2D any-point trajectories and the executable 3D robot actions, functioning as an inverse dynamic model and eliminating the necessity of other task specification. However, how to effectively condition the policy using the predicted 2D trajectories remains an unresolved challenge. Based on our analysis, we present two key insights: \textbf{(\romannumeral 1)} Explicit spatial alignment and fusion of the 2D trajectories with image observations turns out to be highly beneficial. It simplifies the mapping of 2D trajectories to image space and highlights motion-relevant regions, enabling the policy model to focus on important areas and reducing the learning difficulty. \textbf{(\romannumeral 2)} Trajectory positions exhibit different characteristics: starting points typically emphasize local motion, while endpoints focus more on global trend. Therefore, an adaptive trajectory representation and conditioning technique is expected to flexibly guide the policy by effectively capturing these variations.

Based on the aforementioned analysis, we propose an adaptive policy conditioning technique to incorporate trajectory information into 2D images. Specifically, we construct an additional mask modality, and populate it with 2D trajectories based on their specific spatial locations, while setting the value of each point in the trajectory mask as a learnable embedding. Before encoding the image, we concatenate this mask modality with the image observations on the channel dimension, forming a tensor of dimensions $H \times W \times 4$, as shown in Fig.~\ref{MoE}(b). To sum up, our adaptive policy conditioning technique yield an adaptive 2D trajectory representation, which could be explicitly aligned with the image observations to guide the policy prediction more flexibly.





\subsection{Training}
\label{Training}


We argue that trajectory models could be be shared by multiple domains data. Therefore, we first pre-train a generalizable trajectory model on $\boldsymbol{\mathcal{D}_{ood}} \boldsymbol{\cup} \boldsymbol{\mathcal{D}_{in}}$. The $\boldsymbol{\mathcal{D}_{ood}}$ typically contains additional environments, objects, skills, and embodiments. Then we train the trajectory-guided policy model on $\boldsymbol{\mathcal{D}_{in}}$ based on the pre-trained trajectory model. In trajectory-guided policy training phase, we freeze the trajectory model.

\noindent \textbf{Pre-training sparsely-gated MoE-based trajectory model.} Following ATM, we train the trajectory model using the ground truth tracks
generated by CoTracker~\citep{cotracker}. We use MSE to optimize the trajectory prediction loss $\mathcal{L}_{\operatorname{tra}}$. Additionally, we also consider adding an auxiliary load-balancing loss~\citep{shard} to balance the activation frequency among experts. Given a batch of queries $\mathcal{B}$, it contains B samples and each sample contains S tokens. We define $\mathcal{Q}_i$ as the proportion of tokens distributed to expert $i$, and $\mathcal{P}_i$ as the proportion of the gating probability assigned to expert $i$. The $\mathcal{L}_{\operatorname{lo-ba}}$ can be formulated as follows:

\begin{align}
&\mathcal{Q}_i = \frac{1}{S \cdot B}\sum_{x_{s}^{m} \in \mathcal{B}}\mathbbm{1}\{\operatorname{argmax} \mathcal{G}(\mathbf{x_{s}^{m}}; \mathbf{\Theta}) = i\},\label{eq:token_sum}\\
&\mathcal{P}_i = \frac{1}{S \cdot B }\sum_{x_{s}^{m} \in \mathcal{B}} \mathcal{G}(\mathbf{x_{s}^{m}}; \mathbf{\Theta})_i,\label{eq:prob_sum}\\
&\mathcal{L}_{\operatorname{lo-ba}} = N\sum_{i=1}^{N}\mathcal{Q}_i\mathcal{P}_i.\label{eq:load_balance}
\end{align}

Additionally, ST-MoE~\citep{stmoe} proposed using the router z-loss to penalize large logits entering the gating network, thereby improving training stability. We  similarly applies it to the training of our trajectory model, as shown below:

\begin{equation}
    \mathcal{L}_{z}(x) = \frac{1}{S} \sum_{k=1}^S \left(\log \sum_{i=1}^N e^{\mathsf{g}_i^{(k)}} \right)^2,
\label{eqn: z_loss}
\end{equation}
where $S$ is the total number of tokens, $N$ is the total number of experts, and $\mathsf{g} \in \mathcal{R}^{S \times N}$ are the logits going into the router. In summary, the trajectory model's training loss is as follow:

\begin{equation}
\ \mathcal{L}_{\operatorname{total}} = \lambda_{\operatorname{tra}} \cdot \mathcal{L}_{\operatorname{tra}} + \lambda_{\operatorname{lo-ba}} \cdot \mathcal{L}_{\operatorname{lo-ba}} + \lambda_{z} \cdot \mathcal{L}_{z}.
\end{equation}

\noindent \textbf{Training trajectory-guided policy through adaptive policy conditioning technique.} For trajectory-guided policy training, we employ the paradigm of behavior cloning. Specifically, we use the MSE between predicted actions and ground truths as the loss function.

\section{Experiments}

We perform evaluation experiments using the LIBERO benchmark~\citep{libero} and a real
low-cost dual-arm robot~\citep{robot}. 
In experiments, we quantitatively find that the success rate of downstream manipulation tasks is generally positively correlated with the performance of the trajectory model. Detailed quantitative experiment results can be found in the appendix. Meanwhile, CoTracker~\cite{cotracker} can provide relatively accurate trajectory annotations; therefore, we report the average success rate of downstream manipulation tasks as our metric. 
Our experiments aim to study the following questions:

\noindent \textbf{Q1:} Does the techniques of router z-loss, load-balancing loss, and adding noise to gating logits enhance the training of sparsely-gated Tra-MoE?

\noindent \textbf{Q2:} Does sparsely-gated Tra-MoE, compared to the dense Tra-baseline, benefit more from large-scale out-of-domain data (such as cross-embodiments and different physics engine data) and demonstrate superior scaling up capabilities?



\noindent \textbf{Q3:} When we scale up the dense Tra-baseline to match the parameter count of Tra-MoE, does the sparsely-gated Tra-MoE still maintain its advantage?

\begin{table*}[htbp]
    \centering
    \begin{subtable}[t]{0.49\linewidth}
    \label{ablation1}
     \centering
       {
             \begin{tabular}{cccccc}
	\toprule[0.11em]
                  & Spatial & Goal & Object & Long & Avg. \\ \hline
$0$ & 62.0    & 73.0 & 71.0   & 21.5 & 56.9 \\
$5{\cdot}10^{-4}$ & 69.5    & 73.5 & 68.0   & 22.5 & 58.4 \\
$1{\cdot}10^{-4}$ & 62.5    & 81.0 & 73.5   & 28.5 & \textbf{61.4} \\
$1{\cdot}10^{-5}$ & 60.0    & 65.5 & 68.5   & 27.0 & 55.3 \\
$1{\cdot}10^{-6}$ & 60.0    & 79.0 & 53.0   & 31.5 & 55.9 \\ 
\bottomrule[0.11em]
\end{tabular}

		    }
            \caption{The value of $\lambda_{z}$.}
            \label{ablation1}
    \end{subtable}
    \hfill
    \begin{subtable}[t]{0.49\linewidth}
    \label{ablation2}
    \centering
        {
        \begin{tabular}{cccccc}
        \toprule[0.11em]
                  & Spatial & Goal & Object & Long & Avg. \\ \hline
$0$               & 62.0    & 73.0 & 71.0   & 22.5 & 56.9 \\
$1{\cdot}10^{-3}$ & 58.5    & 74.0 & 60.0   & 18.0 & 52.6 \\
$1{\cdot}10^{-5}$ & 61.0    & 69.5 & 56.0   & 26.0 & 53.1 \\
$1{\cdot}10^{-7}$ & 66.5    & 67.0 & 73.5   & 22.0 & \textbf{57.3} \\ 
\bottomrule[0.11em]
\end{tabular}
		    }
    \caption{The value of $\lambda_{lo-ba}$.}
     \label{ablation2}
    \end{subtable}

    \bigskip
\vspace{-3mm}
    \begin{subtable}[t]{0.49\linewidth}
    \label{ablation3}
    \centering
       {
        \begin{tabular}{cccccc}
\toprule[0.11em]
           & Spatial & Goal & Object & Long & Avg. \\ \hline
w/o. noise & 62.0    & 73.0 & 71.0   & 22.5 & \textbf{56.9} \\
w. noise & 71.5    & 67.0 & 57.5   & 27.0 & 55.8 \\ 
\bottomrule[0.11em]
\end{tabular}
			}
    \caption{Adding a noise term to the gating logits.}
    \label{ablation3}
    \end{subtable}
    \hfill
    \begin{subtable}[t]{0.49\linewidth}
    \label{ablation4}
    \centering
       {
       \begin{tabular}{cccccc}
\toprule[0.11em]
      & Spatial & Goal & Object & Long & Avg. \\ \hline
MoE    & 62.5    & 81.0 & 73.5   & 28.5 & \textbf{61.4}\\
Width & 66.5    & 50.5 & 61.5   & 15.0 & 48.4 \\
Depth & 61.5    & 75.5 & 46.5   & 26.5 & 52.5 \\ 
\bottomrule[0.11em]
\end{tabular}
             }
     \caption{Scaling Tra-baseline from width and depth.}
     \label{ablation4}
    \end{subtable}

    \bigskip
\vspace{-3mm}
    \begin{subtable}[t]{0.49\linewidth}
    \label{ablation5}
    \centering
      {
       \begin{tabular}{cccccc}
\toprule[0.11em]
Num.  & Spatial & Goal & Object & Long & Avg. \\ \hline
1 & 49.5    & 67.0 & 56.5   & 35.0 & 52.0 \\
2 & 59.0    & 72.5 & 65.5   & 27.0 & 56.0 \\
3 & 64.0    & 68.0 & 71.0   & 17.5 & 55.1 \\
4 & 62.0    & 73.0 & 71.0   & 21.5 & \textbf{56.9} \\ 
\bottomrule[0.11em]
\end{tabular}
			}
            \caption{The number of experts.}
            \label{ablation5}
    \end{subtable}
    \hfill
    \begin{subtable}[t]{0.49\linewidth}
    \label{ablation6}
    \centering
     {
     \begin{tabular}{cc|ccccc}
\toprule[0.11em]
MoE & OOD & Spatial & Goal & Object & Long & Avg. \\ \hline
    &          & 67.5    & 68.5 & 68.0   & 26.5 & 57.6 \\ 
\checkmark   &          & 63.0    & 64.0 & 60.5   & 19.5 & 51.8 \\
    & \checkmark       & 49.5    & 67.0 & 56.5   & 35.0 & 52.0 \\
\checkmark   & \checkmark       & 62.5    & 81.0 & 73.5   & 28.5 & \textbf{61.4} \\ 
\bottomrule[0.11em]
\end{tabular}
             }
            \caption{Scaling with MoE and OOD data.}
     \label{ablation6}
    \end{subtable}
    \vspace{-3mm}
 \caption{The ablation experiments of the sparsely-gated Tra-MoE. All baselines are the track transformer of ATM.}
 \vspace{-3mm}
\label{tab:array}
\end{table*}

\noindent \textbf{Q4:} Is our proposed adaptive policy conditioning technique for trajectory-guided policy more effective than ATM?


\subsection{Simulation Experiments}

\noindent \textbf{Experiment setup.} The LIBERO~\citep{libero} is divided into five categories: LIBERO-Spatial, LIBERO-Object, LIBERO-Goal, LIBERO-Long, and LIBERO-90. LIBERO-90 includes 90 tasks, while each of the other categories contains 10 tasks. Compared to other categories, LIBERO-90 presents a highly varied selection of object categories, layouts, environments, and skill goals. It is typically used for \textit{pre-training} rather than  \textit{joint co-training}. Therefore, our main ablation experiments select 90 tasks from LIBERO-90 to construct dataset $\boldsymbol{\mathcal{D}_{ood}}$, utilizing 20 action-free videos per task; and choose 40 tasks from LIBERO-Spatial, LIBERO-Object, LIBERO-Goal, and LIBERO-Long to build dataset $\boldsymbol{\mathcal{D}_{in}}$, using 10 action-labeled demonstrations per task. In summary, for training Tra-MoE, our main ablation experiments use a total of $2,200$ action-free videos, with an out-of-domain to in-domain data ratio of $9:2$. Additionally, on this basis, we also further select 92 tasks from the RLbench~\cite{rlbench} (using CoppeliaSim rather than MuJoCo as physics engine) simulation environment and generate 5 videos for each task, totaling $2,660$ videos in conjunction with the LIBERO data, to train Tra-MoE. The different physics engines and dynamics properties further increase the data distribution discrepancy in $\boldsymbol{\mathcal{D}_{ood}}$. Finally, as for the trajectory-guided policy, we train a separate multi-task policy for each category. Each policy undergoes behavioral cloning training with 100 action-labeled demonstrations from 10 tasks.


\noindent \textbf{Implement details.} Building on ATM~\citep{atm}, we develop our sparsely-gated Tra-MoE and adaptive policy conditioning technique. Specifically, when using only LIBERO data, we replace the 1th, 5th, and 8th layers of the track transformer with MoE blocks. When further incorporating RLbench data, we additionally replace the 2th and 7th layers. Unless otherwise specified, we set the number of experts to 4 by default and use a \textbf{top-1} gating strategy for token choice to ensure the FLOPs per token remain constant.





\noindent \textbf{Experiment results and analysis.} We present empirical simulation results to address the aforementioned research questions and make the following findings:

\noindent \textcolor{blue}{\textbf{\textit{Result 1:}}} In Tab.~\ref{ablation1}, we examine the effect of the router z-loss weight $\lambda_{z}$. The results indicate that a higher $\lambda_{z}$ can effectively enhance performance, with an improvement of 4.5 (61.4 vs. 56.9) at $\lambda_{z} =1 \cdot 10^{-4}$. In Tab.~\ref{ablation2}, we study the impact of the load-balancing loss weight $\lambda_{lo-ba}$. The results show a significant performance decline with larger values of $\lambda_{lo-ba}$, with only a slight improvement (56.7 → 57.3) at $\lambda_{lo-ba} =1 \cdot 10^{-7}$. In Tab.~\ref{ablation3}, we verify the effectiveness of adding noise to the gating logits, which results in a performance decrease of by 1.1 (56.9 → 55.8).

\noindent \textcolor{blue}{\textbf{\textit{Finding 1:}}} Based on the above results, we find that using router z-loss to penalize large logits entering the gating network can effectively improve training stability and lead to performance gains. Meanwhile, the load-balancing loss used to balance the activation frequency among experts and the addition of noise to enhance expert allocation exploration cannot bring effective performance improvements. An intuitive explanation is that when the data distribution is uneven, the load-balancing loss might force experts to set shared parameters on data with a large domain gap, which could lead to mutual counteraction of learning gradients. This undermines the advantage of parameter specialization in sparsely-gated MoE. Similarly, adding noise is also detrimental to learning parameter specialization among experts.


\begin{table}[t]\centering
\label{adaptive}
\begin{tabular}{lccccc}
\toprule[0.11em]
 &Spatial &Goal &Object &Long &Avg. \\ \hline
ATM &51.0 &42.0 &55.5 &19.0 &41.9\\
Tra-MoE &54.0 &62.0 &61.0 & 41.0&\textbf{54.5}\\
\bottomrule[0.11em]
\end{tabular}
\caption{The comparison results when training trajectory model using RLbench video data.}
\label{rlbench}
\vspace{-6mm}
\end{table}

\begin{table}[t]\centering
\label{adaptive}
\begin{tabular}{lccccc}
\toprule[0.11em]
 &Spatial &Goal &Object &Long &Avg. \\ \hline
ATM &62.5 &81.0 &73.5 &28.5 &61.4 \\
\textbf{+} DM  &69.0 &58.0 &85.5 & 33.5&61.5\\

\textbf{+} AM &69.5 &77.0 &88.0 &30.5 & \textbf{66.3}\\


\bottomrule[0.11em]
\end{tabular}
\caption{The ablation experiments of the adaptive policy conditioning technique. DM (RT-Trajectory style) and AM respectively represent hand-drawn mask and adaptive mask.}
\label{adaptive}
\vspace{-6mm}
\end{table}

\begin{table*}[t]\centering
\label{reals}
\begin{tabular}{ccc|cccccc}\toprule[0.11em]
Human Data &MoE &Adaptive Trajectory Mask  &Pour &Push &Pick and Pass &Tissue &Fold  &Average\\ \hline
 &  & &40.0 &45.0  & 50.0& 30.0&25.0 &38.0\\
 \checkmark&  & & 45.0& 35.0 & 50.0&25.0 &35.0 &38.0\\

 \checkmark  &\checkmark & &40.0 & 50.0&60.0 &35.0 &45.0 &46.0\\

 \checkmark &\checkmark &\checkmark &60.0 &70.0 &65.0 &35.0 &50.0 &\textbf{56.0} \\

\bottomrule[0.11em]
\end{tabular}
\caption{The ablation of ood human data, MoE architecture, and adaptive policy conditioning technique on our real-world tasks.}
\vspace{-4mm}
\label{reals}
\end{table*}

\noindent \textcolor{blue}{\textbf{\textit{Result 2:}}} We comprehensively investigate the impact of introducing sparsely-gated MoE and large-scale out-of-domain data on scaling up trajectory models in Tab.~\ref{ablation6}. Firstly, when training only with in-domain data, Tra-baseline achieves an average success rate of 57.6, while Tra-MoE only achieves an average success rate of 51.8. Secondly, when adding out-of-domain data to jointly train Tra-baseline, the average success rate drops to 52.0 (57.6 → 52.0), with particularly significant performance declines in LIBERO-Spatial and  LIBERO-Object. However, when adding out-of-domain data to jointly train Tra-MoE, the average success rate increases to 61.4 (51.8 → 61.4). Additionally, in Tab.~\ref{ablation5}, we also explore the impact of the number of experts in sparsely-gated MoE. The results show that using just two experts can lead to an increase of 4.0 (52.0 vs. 56.0) in the average success rate. With additional experts, performance generally trends upward, albeit with some fluctuations. Finally, we report the comparison results of training trajectory model  by further using RLbench data in Tab.~\ref{rlbench}. The results indicate that although the introduction of RLbench data led to a performance drop in LIBERO, our Tra-MoE still outperform the baseline by 12.6 (54.5 vs. 41.9).


\noindent \textcolor{blue}{\textbf{\textit{Finding 2:}}} Based on the above results, we have the following findings and analysis: \textcolor{blue}{\textbf{(\romannumeral 1)}} When the data scale is small, naively expanding the model capacity through sparsely-gated MoE does not lead to performance improvements. A explanation is that the larger model may lead to a overfitting problem when training on small-scale data. Moreover, small-scale data often leads to some experts not being sufficiently trained, thereby affecting performance. \textcolor{blue}{\textbf{(\romannumeral 2)}} Naively introducing out-of-domain data for joint training often leads to performance degradation in the target domain. However, simultaneously incorporating out-of-domain data and expanding models with sparsely-gated MoE may improve performance or alleviate performance degradation. This phenomenon suggests that although out-of-domain data generally enhances model generalization, it could hinder task-specific performance in the target domain due to insufficient in-domain data learning. In contrast, expanding the model with out-of-domain data using sparsely-gated MoE achieves a more optimal balance between parameter cooperation and specialization across diverse data and even different tokens within the same data input. This design allows these large-scale data to jointly train most of parameters, ensuring the capture of general patterns for mutual cooperation. \textcolor{blue}{\textbf{(\romannumeral 3)}} Even with a few experts, Tra-MoE exhibits significant scaling up capabilities. Additionally, further increasing the number of experts generally leads to performance improvement. However, this improvement is not guaranteed due to factors like data-expert complexity matching and training stability.


\begin{figure*}[htbp]
\label{demo}
\centering
\includegraphics[width=0.85\linewidth]{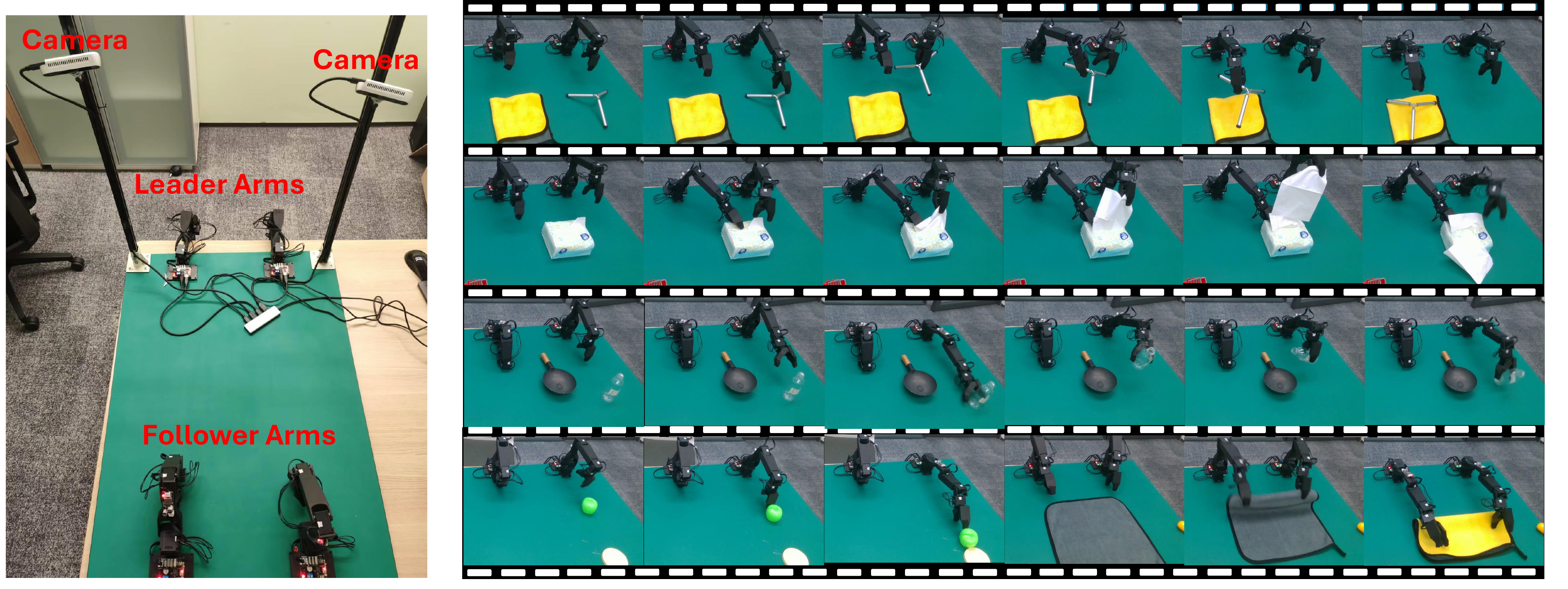}
\caption{\textbf{Left:} The real-world experiments hardware platform setup. \textbf{Right:} The real-world tasks evaluation demonstrations.}
\vspace{-5mm}
\label{demo}
\end{figure*}

\noindent \textcolor{blue}{\textbf{\textit{Result 3:}}} In Tab.~\ref{ablation4}, we also investigate the performance comparison between the dense Tra-baseline and the sparse Tra-MoE with the same number of parameters. Specifically, we expand Tra-baseline to match the model capacity of Tra-MoE by increasing its model width or depth separately. These expanded versions achieve average success rates of 48.4 and 52.5 respectively, which are significantly lower than Tra-MoE's 61.4.

\noindent \textcolor{blue}{\textbf{\textit{Finding 3:}}} Based on these results, we are surprised to find that simply expanding the dense Tra-baseline in depth and width can not effectively improve performance. This highlights the importance of the sparsely-gated MoE architecture when training with large-scale out-of-domain data, as it ensures dynamic activation of different experts based on different data input and different tokens within the same data input, significantly reducing optimization conflict problems.

\noindent \textcolor{blue}{\textbf{\textit{Result 4:}}} In Tab.~\ref{adaptive}, we investigate the effectiveness of the adaptive policy conditioning. Specifically, building upon ATM, we first introduce an additional hand-drawn mask modality before image encoding. In this mask, the first half of the trajectory is set to $128$, while the second half is set to $255$. We call it `ATM + hand-drawn mask' and this is similar to RT-Trajectory~\cite{rt-tra}. Next, we replace the hand-drawn mask with an adaptive mask, where each point on the trajectory is instantiated as a learnable embedding. We call it `ATM + adaptive mask'. The results show that `ATM + hand-drawn mask' achieves only a 0.1 improvement compared to the baseline. However, except for a significant performance decrease in LIBERO-Goal, it achieves performance improvements of 6.5, 12.0, and 5.0 in LIBERO-Spatial, LIBERO-Object, and LIBERO-Long respectively. In addition, the `ATM + adaptive mask' achieves a performance improvement of 4.9 (66.3 vs. 61.4) compared to the baseline. It further improves performance across all suites compared to `ATM + hand-drawn mask', especially achieving a 19.0 performance gains in LIBERO-Goal (77.0 vs. 58.0).

\noindent \textcolor{blue}{\textbf{\textit{Finding 4:}}} Based on these results, we find that explicitly mapping 2D trajectories to visual observations significantly improves the performance of LIBERO-Spatial, LIBERO-Object, and LIBERO-Long, but causes a notable decrease in LIBERO-Goal. By instantiating the points in the 2D trajectory as learnable embeddings, we not only further enhances the performance of LIBERO-Spatial, LIBERO-Object, and LIBERO-Long, but also significantly improves LIBERO-Goal. We analyze that the decline in LIBERO-Goal is mainly due to overlapping front and back parts of some trajectories in the 2D images, leading to mapping errors. However, our adaptive trajectory representation and conditioning technique effectively mitigates this issue. In summary, these experiments collectively demonstrate the effectiveness of explicitly aligning 2D trajectories with images in space as well as applying the adaptive conditioning technique.





\subsection{Real-World Experiments}


\noindent \textbf{Experiment setup.} As shown in Fig~\ref{reals}, we utilize a dual-arm robot~\cite{robot} to conduct real-world experiments.  We evaluate 5 tasks, which include two single-arm tasks: pouring water and pushing a vegetable to the side of the cutting board, as well as three dual-arm tasks: pulling out tissues, folding towels, and picking up and passing a holder. Specifically, for each task, we collect 50 action-labeled robot demonstrations. Concurrently, we also collect 50 videos of human (cross-embodiments) performing for each task. Compared to the robot data, the human video encompass more environment variations and distractor objects. Please see the appendix for more detailed evaluation details and real-world setup.



\noindent \textbf{Implement details.} Following our simulation experiments, we utilize two camera views and uniformly resize their resolution to 128×128 for trajectory model and policy training. We train our real-world Tra-MoE on borh human video and robot video data, with a separate policy trained for each task. Therefore, our real-world Tra-MoE integrates data across different embodiments (human, single-arm robot, and dual-arm robot). They have significantly different visual appearances and dynamics properties. Finally, we also present some real-world task evaluation demonstrations in Fig.~\ref{demo}.

\noindent \textbf{Experiment results and analysis.} To further strengthen our conclusions that sparsely-gated MoE can better learn trajectory model from out-of-domain data and our adaptive policy condition technique could lead to more flexible guidance for policy learning, we conduct real-world ablation experiments, as shown in Tab.~\ref{reals}. The results indicate that simply introducing human video data cannot lead to performance improvement. However, when expanding the model by MoE architecture, the performance increases from 38\% to 46\%. We speculate this may be due to the significant differences in appearance and dynamics between human and robot, with the former typically moving much faster and more flexibly. Finally, when further incorporating an adaptive trajectory mask for policy conditioning, the performance is further increased from 46\% to 56\%.



\vspace{-3mm}

\section{Conclusion}
\label{conclusion}

In this work, we have introduced \textbf{Tra-MoE}, a novel framework that utilizes a sparsely-gated MoE architecture with \textbf{Top-1} gating strategy to learn trajectory model from multiple domains. Tra-MoE achieves superior parameter cooperation and specialization, enabling more effective utilization of large-scale out-of-domain data while maintaining constant FLOPs per token. We train our Tra-MoE and conduct a comprehensive empirical study, demonstrating that Tra-MoE consistently exhibits superior performance compared to the dense baseline, even when the latter is scaled to match Tra-MoE’s parameter count. Additionally, we propose an adaptive policy conditioning technique for trajectory-guided policy, ensuring adaptive 2D trajectory representation and explicit spatial alignment with observations, thereby achieving superior performance. Extensive experiments in both simulated and real-world environments validate our findings and demonstrate the effectiveness of our proposed methods. 


\noindent \textbf{Acknowledgements}

This work is supported by the National Key R$\&$D Program of China (No. 2022ZD0160900), Jiangsu Frontier Technology Research and Development Program (No. BF2024076), and the Collaborative Innovation Center of Novel Software Technology and Industrialization.

{
    \small
    \bibliographystyle{ieeenat_fullname}
    \bibliography{main}
}
\clearpage
\setcounter{page}{1}
\maketitlesupplementary

\appendix

\renewcommand\thesection{\Alph{section}} 

\section{Appendix}

\subsection{Limitations and Discussion}

In our work, we firstly demonstrate the effectiveness of using sparsely-gated Mixture-of-Experts (MoE) for learning trajectory prediction model from large-scale out-of-domain data, while
conducting a thorough experimental study of its training techniques. To the best of our knowledge,
this is one of the first attempts of using sparse MoE architecture for learning from multiple domains data to ensure better parameter cooperation and specialization in robotics. We hope it can serve as a strong baseline and facilitate further research in this direction. While we are encouraged by the strong results across a wide range of simulated and real-world experiments, some limitations and future works still remain. On the one hand, the trajectory prediction model learning can further integrate larger-scale human and robot video data. On the other hand, our adaptive policy condition technique can also be extended to other visual prompts.


\subsection{The stability of LIBERO experiment results}

In our LIBERO simulation experiment results, the training process and results are completely reproducible. Due to the slight randomness in the simulation rendering, the downstream LIBERO evaluation results cannot be fully reproduced. Therefore, we further run the key experiments three times and report the mean and standard deviation in Tab.~\ref{rerun}. The results indicate that the improvements brought by our Tra-MoE and adaptive policy conditioning technique are \textbf{significant and stable}.


\begin{table*}[]
\centering
\begin{tabular}{ccccccl}
\hline
                                        & Spatial & Goal & Object & Long & Avg. &                            \\
                                        \hline
\multirow{3}{*}{Tra-baseline}           & 49.5    & 67.0 & 56.5   & 35.0 & 52.0 & \multirow{3}{*}{\textbf{52.0±0.1}} \\
                                        & 51.5    & 69.0 & 58.5   & 29.0 & 52.0 &                            \\
                                        & 51.5    & 67.0 & 58.5   & 31.5 & 52.1 &                            \\ \hline
\multirow{3}{*}{Tra-MoE}                & 62.5    & 81.0 & 73.5   & 28.5 & 61.4 & \multirow{3}{*}{\textbf{61.2±0.4}} \\
                                        & 62.5    & 80.5 & 72.5   & 27.5 & 60.8 &                            \\
                                        & 67.0    & 79.5 & 72.5   & 27.0 & 61.5 &                            \\ \hline
\multirow{3}{*}{Tra-MoE + Aaptive Mask} & 72.5    & 74.0 & 86.5   & 34.5 & 66.8 & \multirow{3}{*}{\textbf{67.7±0.8}} \\
                                        & 73.0    & 78.0 & 86.5   & 35.5 & 68.3 &                            \\
                                        & 72.5    & 75.5 & 87.5   & 36.0 & 67.9 &                            \\ \hline
\end{tabular}
\caption{We rerun the key experiments three times and report the mean and standard deviation.}
\label{rerun}
\end{table*}

\begin{table*}[ht]
\centering
\begin{tabular}{@{}ccc@{}}
    \toprule
    Hyperparameters & In-domain data & Out-of-domain data \\
    \midrule
    Number of videos & 400 & 2200 / 2660 \\
    Number of tasks & 40 & 130 / 222 \\
    Epoch & 1000 & 300 \\
    Batch size & \multicolumn{2}{c}{2048} \\
    Optimizer & \multicolumn{2}{c}{AdamW} \\
    Learning rate & \multicolumn{2}{c}{1e-4} \\
    Weight decay & \multicolumn{2}{c}{1e-4} \\
    LR scheduler & \multicolumn{2}{c}{Cosine} \\
    LR warm-up & \multicolumn{2}{c}{5} \\
    Clip grad & \multicolumn{2}{c}{10} \\
    Point sampling & \multicolumn{2}{c}{Variance filtering} \\
    Number of points & \multicolumn{2}{c}{32} \\
    Track length & \multicolumn{2}{c}{16} \\
    Augmentation & \multicolumn{2}{c}{ColorJitter, RandomShift} \\
     dropout & \multicolumn{2}{c}{0.2} \\
     depth & \multicolumn{2}{c}{8} \\
     dimension & \multicolumn{2}{c}{384} \\
    \bottomrule
\end{tabular}
\caption{Hyperparameters of our trajectory model training.}
\label{trajectory}
\end{table*}

\begin{table*}[ht]
\makeatletter\def\@captype{table}
\centering
  \begin{tabular}{@{}cc@{}}
    \toprule
    Hyperparameters & Policy  \\
    \midrule
    Number of demonstrations &100 \\
    epoch & 120  \\
    batch size &  384  \\
    optimizer &  AdamW  \\
    learning rate & 5e-4  \\
    weight decay & 1e-4 \\
    lr scheduler & Cosine  \\
    lr warm up & 0 \\
    clip grad & 100 \\
    point sampling & grid \\
    number of points & 32 \\
    track length & 16 \\
    frame stack & 10 \\
    augmentation & ColorJitter,RandomShift \\
     dropout & 0.1 \\
    \bottomrule
  \end{tabular}
 \caption{Hyperparameters of our trajectory-guided policy training.}
  \label{policy}
\end{table*}

\subsection{The comparison of trajectory prediction model}

\begin{figure*}[htbp]
\centering
\includegraphics[width=0.9\textwidth]{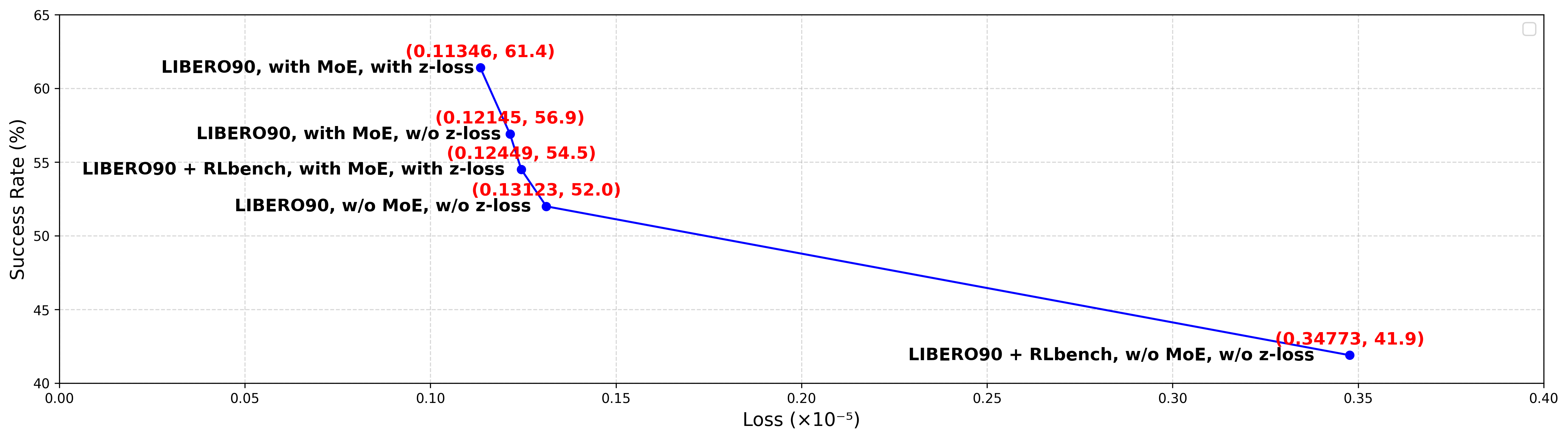}
\caption{The quantitative relationship between downstream policy success rate and trajectory model performance.} 
\label{vis2}
\end{figure*}

In our work, we find that CoTracker~\cite{cotracker} can provide highly accurate labels. On the one hand, we can directly use the ground truth provided by CoTracker to calculate the MSE error with our prediction results to measure the performance of trajectory prediction. On the other hand, considering that trajectory prediction task is multi-modal, we also need directly visualize some samples for prediction performance analysis. \textit{In our experiments, we measure trajectory prediction performance using the MSE loss on validation set of four evaluation suites, as well as qualitatively and quantitatively find that the success rate of downstream manipulation tasks is generally positively correlated with the performance of the trajectory model, as shown in Fig.~\ref{vis2}. Therefore, we report the average success rate of downstream manipulation tasks as our metric.}

We also fairly compare the Tra-baseline (the track transformer of ATM) and our Tra-MoE, which are both trained with a mixture of LIBERO and RLbench video data. Their respective MSE error on the LIBERO validation set are \textbf{0.0000034773} and \textbf{0.0000012449}. Additionally, we further randomly select some samples for visualization, as shown in Fig.~\ref{vis}. The visualization results also indicate that Tra-MoE is significantly more accurate than Tra-baseline. Tra-MoE is generally able to predict accurate trajectories, whereas Tra-baseline occasionally predict stationary or even opposite-direction movements, leading to poorer downstream policy. This is primarily attributed to optimization conflicts arising from multiple domains data joint training. Conversely, our Tra-MoE, with its superior parameter cooperation and specialization, can better handle such situations.

\begin{figure*}[htbp]
\centering
\includegraphics[width=0.99\textwidth]{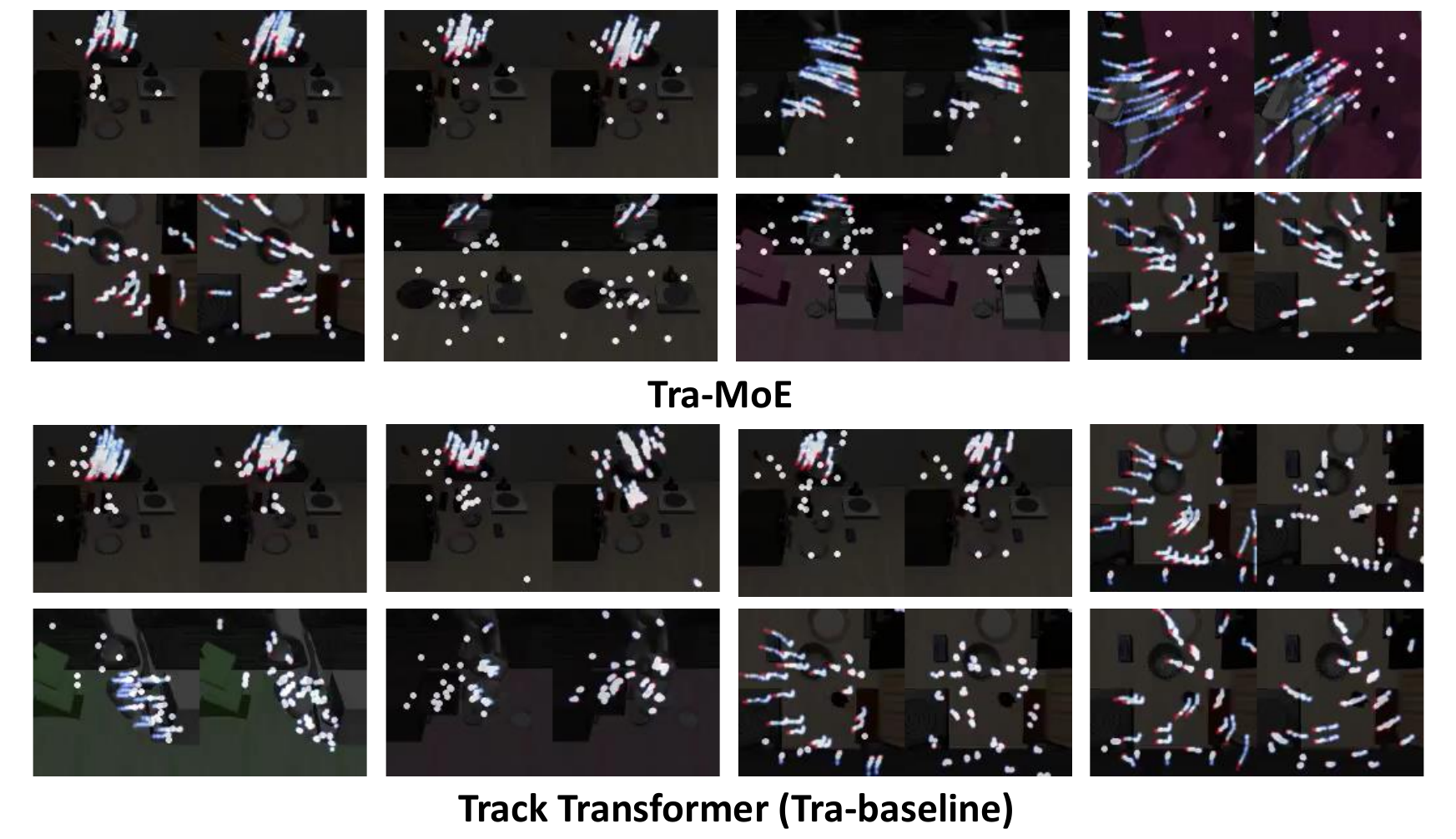} 
\label{seed}
\caption{The trajectory prediction results visualization. \textbf{Left:} the ground truth by CoTracker~\cite{cotracker}. \textbf{Right:} the prediction result.}
\label{vis}
\vspace{-4mm}
\end{figure*}

\subsection{The Simulation environments details}

In this section, we further elaborate on the details of our simulation experiments. The training hyperparameters for the trajectory prediction model and the trajectory-guided policy are shown in Tab.~\ref{trajectory} and Tab.~\ref{policy}, respectively. We use the same training hyperparameters to ensure a fair comparison between our Tra-MoE and Tra-baseline. When we train the trajectory model integrating RLbench data, we report the specific tasks used in Tab.~\ref{rlbench}. For the majority of the hyperparameters, we inherit the settings from ATM~\citep{atm}. Additionally, when we extend Tra-baseline in depth, the depth is increased from 8 to 14; when we extend Tra-baseline in width, the dimension is increased from 384 to 512. Finally, following the original LIBERO~\citep{libero} setup, we perform 20 trials for each task evaluation, ensuring a total of 800 (20×40) trials for each model evaluation.

\subsection{The Real-World environments details}

In this section, we further elaborate on the details of our real-world experiments. Specifically, we use two leader arms to perform teleoperation for follower arms data collection, where 50 demonstrations are collected for each task for trajectory model and policy training. For our real-world evaluations, we conduct 20 trials for each task, while ensuring, to the extent possible, that the object poses in the training set differ from those in the test set. For the relevant training hyperparameters, we maintain consistency with the simulation experiments.

\begin{table*}[h]
\centering
\begin{tabular}{|c|c|}
\hline
beat the buzz &lamp off \\
\hline
block pyramid &place hanger on rack \\
\hline
put umbrella in umbrella stand &take money out safe \\
\hline
place shape in shape sorter &take umbrella out of umbrella stand \\
\hline
reach and drag &take tray out of oven \\
\hline
lamp on &push button \\
\hline
change channel &take toilet roll off stand \\
\hline
light bulb in &setup checkers \\
\hline
play jenga &close door \\
\hline
reach target &open door \\
\hline
take plate off colored dish rack &meat on grill \\
\hline
change clock &close drawer \\
\hline
light bulb out &stack cups \\
\hline
plug charger in power supply &take usb out of computer \\
\hline
remove cups &slide cabinet open and place cups \\
\hline
take shoes out of box &slide block to target \\
\hline
close box &put bottle in fridge \\
\hline
meat off grill &toilet seat down \\
\hline
pour from cup to cup &put groceries in cupboard \\
\hline
scoop with spatula &toilet seat up \\
\hline
press switch &put item in drawer \\
\hline
screw nail &stack blocks \\
\hline
move hanger  &close grill \\
\hline
close fridge &open microwave \\
\hline
open box &put books on bookshelf \\
\hline
setup chess &put knife in knife block \\
\hline
close jar &empty container \\
\hline
open drawer &turn tap \\
\hline
close laptop lid &open grill \\
\hline
open fridge &close microwave \\
\hline
solve puzzle &turn oven on \\
\hline
tv on &wipe desk \\
\hline
put knife on chopping board &hockey \\
\hline
stack wine &take cup out from cabinet \\
\hline
unplug charger &put rubbish in bin \\
\hline
get ice from fridge &open wine bottle \\
\hline
open oven &hit ball with queue \\
\hline
put money in safe &weighing scales \\
\hline
straighten rope &sweep to dustpan \\
\hline
water plants &put plate in colored dish rack \\
\hline
hang frame on hanger &open window \\
\hline
phone on base &put tray in oven \\
\hline
put shoes in box &place cups \\
\hline
take frame off hanger &insert usb in computer \\
\hline
insert onto square peg &take item out of drawer \\
\hline
pick and lift put toilet roll on stand &take lid off saucepan\\
\hline
\end{tabular}
\caption{The language annotations of 92 RLbench tasks.}
\label{rlbench}
\end{table*}


\end{document}